\documentclass[10pt,twocolumn,letterpaper]{article}

\usepackage{cvpr}
\usepackage{times}
\usepackage{epsfig}
\usepackage{graphicx}
\usepackage{amsmath}
\usepackage{amssymb}
\usepackage{bm}
\usepackage{multirow}
\usepackage{diagbox}
\usepackage{booktabs}
\usepackage{authblk}

\def\etal{\textit{et~al.}}

\def\ie{\textit{i.e.}}

\usepackage[pagebackref=true,breaklinks=true,letterpaper=true,colorlinks,bookmarks=false]{hyperref}

 \cvprfinalcopy 

\def\cvprPaperID{410} 

\ifcvprfinal\fi
\begin{document}

\title{Linkage Based Face Clustering via Graph Convolution Network}




\author{%
\vspace{-0.6cm}
Zhongdao Wang$^1$, Liang Zheng$^2$, Yali Li$^1$, Shengjin Wang$^1$\thanks{Corresponding author}\\
{$^1$Department of Electronic Engineering, Tsinghua University}\\
{$^2$Australian Centre for Robotic Vision, Australian National University}\\
{\texttt{\small{wcd17@mails.tsinghua.edu.cn}} \hspace{0.5cm}}
{\texttt{\small{liang.zheng@anu.edu.au}} \hspace{0.5cm}}
{\texttt{\small{\{liyali13, wgsgj\}@tsinghua.edu.cn}}}
\vspace{-0.5cm}
}

\maketitle

\begin{abstract}
In this paper, we present an accurate and scalable approach 
to the face clustering task. We aim at grouping a set of faces by their potential identities. 
We formulate this task as a link prediction problem: a link exists between two faces if they are of the same identity.
The key idea is that we find the local context in the feature space around an instance (face) contains rich information about the linkage relationship between this instance and its neighbors. 
By constructing sub-graphs around each instance as input data, which depict the local context, we utilize the graph convolution network (GCN) to perform reasoning and infer the likelihood of linkage between pairs in the sub-graphs.
Experiments show that our method is more robust to the complex distribution of faces than conventional methods, yielding favorably comparable results to state-of-the-art methods on standard face clustering benchmarks, and is scalable to large datasets.
Furthermore, we show that the proposed method does not need the number of clusters as prior, is aware of noises and outliers, and can be extended to a multi-view version for more accurate clustering accuracy. \footnote{Code available at \url{https://github.com/Zhongdao/gcn_clustering/}.}
\end{abstract}
\vspace{-0.5cm}
\section{Introduction}
\label{introduction}

In this paper, we study the problem of clustering faces according to their underlying identities. We assume no prior of the distribution of face representations or the number of identities. 
Face clustering is a fundamental task in face analysis and has been extensively studied in previous works~\cite{rankorder, aro, ddc, con}.
Some key applications include: grouping and tagging faces in desktop / online photo albums for photo management~\cite{rankorder}, organizing large scale face image / video collections for fast retrieval in time-sensitive scenarios like forensic investigations~\cite{boston}, and automatically data cleaning / labeling for constructing large-scale datasets~\cite{mf2, ms1m, cdp}. 
\begin{figure}
    \centering
    \includegraphics[width=0.95\linewidth]{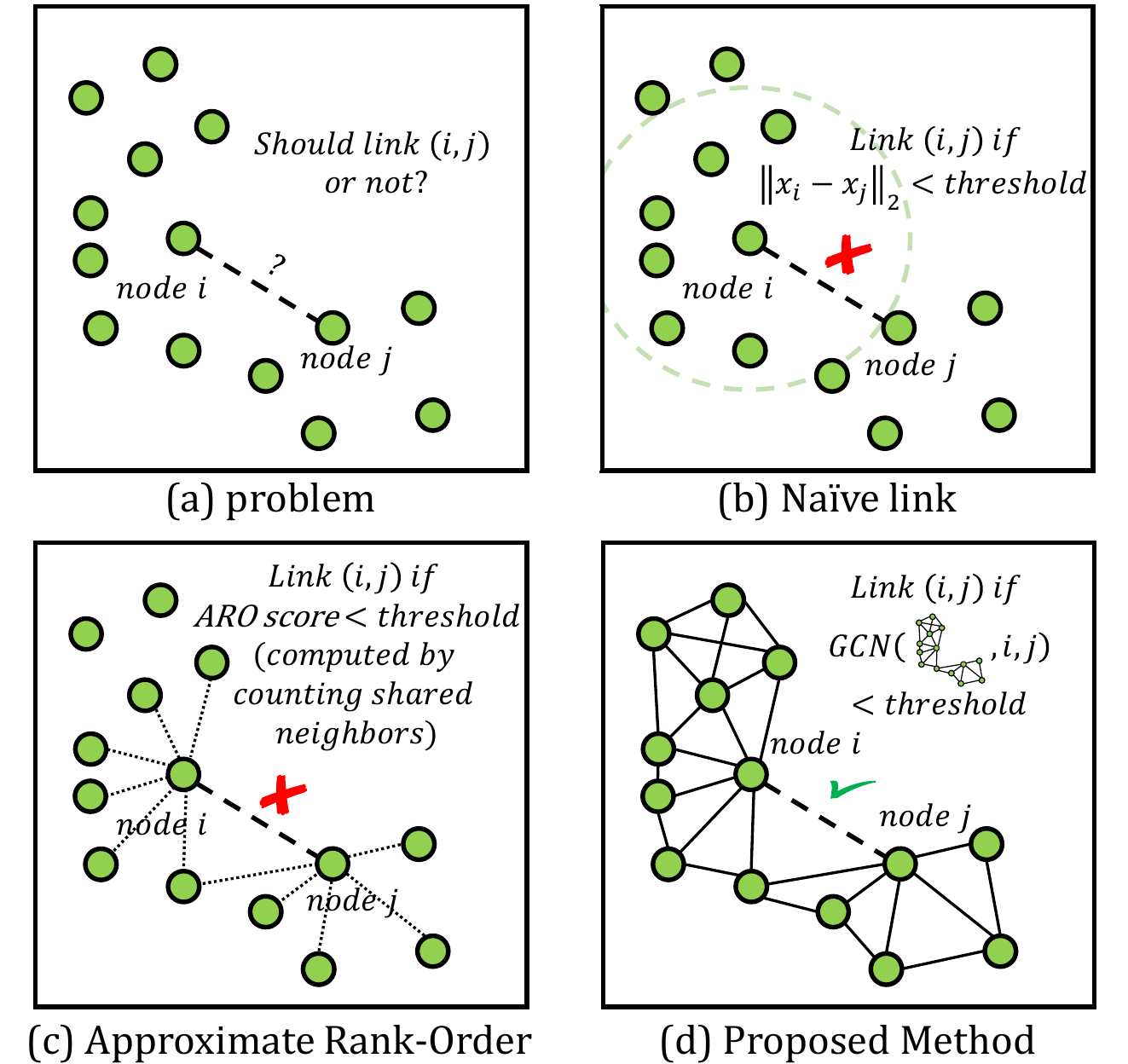}
    \caption{High-level idea of our method. (a): This paper aims to estimate whether two nodes should be linked. (b-d): Comparison of  three linkage estimation methods. (b): Directly thresholding the $l_2$ distance, \emph{without considering context}. (c): Using a heuristic mthod for linkage estimation \emph{based on context}. (d): Our method, \ie, learning the linkage likelihood with a parametric model, which is \emph{context-based}.}
    \label{fig:intro}
    \vspace{-0.5cm}
\end{figure}

Conventional clustering methods suffer from the complex distribution of face representations, because they make impractical assumptions on data distribution. 
For instance,  K-Means~\cite{kmeans} requires the clusters to be convex-shaped, Spectral Clustering~\cite{spectral1} needs different clusters to be balanced in the number of instances, and DBSCAN~\cite{dbscan} assumes different clusters to be in the same density.
In contrast, a family of linkage-based clustering methods make no assumption on data distribution and achieve higher accuracy. 
As shown in Fig.\ref{fig:intro} (a), linkage-based methods aim at predicting whether two nodes (or clusters) should be linked (are of the same identity).
Fig.\ref{fig:intro} (b) presents a naive solution to this problem by directly linking node pairs whose $l_2$ distance is under a certain threshold. 
This is apparently not a good solution since the densities of clusters vary a lot.
Therefore, more sophisticated metrics are designed to compute the linkage likelihood, such as the Approximate Rank-Order Distance (see Fig.\ref{fig:intro} (c)). 

Instead of computing the linkage likelihood by heuristic metrics, we propose to \emph{learn} to predict if two nodes should be linked.
As the key idea of this work, we find that the linkage likelihood between a node and its neighbors can be inferred from its \emph{context}.
In order to leverage the valuable information in the context of nodes, we propose a learnable clustering method based on the graph convolution network (GCN), and the main idea is shown in Fig.\ref{fig:intro} (d).   
The framework of the proposed method can be summarized as follow.

Firstly, we formulate clustering as a link prediction problem~\cite{linkprediction}. That is, a link exists between two nodes when their identity labels are identical. 
Secondly, we only predict linkages between an instance and its nearest neighbors.
Accordingly, we construct an Instance Pivot Subgraph around each instance (pivot), to depict the local context, with each node modeling a pivot-neighbor pair.
From IPS, we can perform reasoning to infer which pivot-neighbor pair should be linked and we adopt GCN to learn this task.
Finally,  GCN outputs a set of linkage likelihood, and we transitively merge linked nodes to obtain the clusters.

We show that the proposed method is accurate when compared with the state-of-the-art methods and is scalable in terms of computational complexity.
The proposed method learns to generate linkage likelihood automatically, and results in superior performance to other linkage-based  methods like ARO \cite{aro} in which the linkage likelihood is computed by heuristic rules.
In addition, our approach is aware of noises and outliers, does not need the number of clusters as input, and is easy to be extended to  a multi-view version for utilizing data from different sources.

The  remainder  of  this  paper  is  organized  as  follows. First,  we briefly  review  the  related  works in  Section  \ref{relatedwork}. Then,  Section \ref{method} introduces the proposed clustering algorithm. The experimental results are presented in Section \ref{experiments} and conclusions are given in Section \ref{conclusion}.

\section{Related Work}
\label{relatedwork}

\textbf{Face Clustering.} Due to large variations in pose, occlusion, illumination and number of instances, face clusters vary significantly in size, shape and density. 
The complex distribution of face representations makes it unsuitable to apply classic clustering algorithms like K-Means~\cite{kmeans} and spectral clustering~\cite{spectral1}, because these methods have rigid assumptions on data distribution.
Several papers develop clustering algorithms based on Agglomerative Hierarchical Clustering (AHC)~\cite{rankorder, ddc, pahc}, which is robust in grouping data with complex distribution. Lin~\etal~\cite{pahc} propose the proximity-aware hierarchical clustering (PAHC) which exploits a linear SVM to classify local positive instances and local negative instances. Zhu~\etal~\cite{rankorder} propose a cluster-level affinity named Rank-Order distance to replace the original $l_1/l_2$ distance, and demonstrate its ability in filtering out noise / outliers. Lin~\etal~\cite{ddc} also design a density-aware cluster-level affinity using SVDD~\cite{svdd} to deal with density-unbalanced data.
All the above methods yield good performance on unconstrained face clustering, but their computation complexity remains a problem, limiting their application in large-scale clustering.

Fortunately, the Approximate Rank-Order Clustering (ARO)~\cite{aro} provides an efficient framework for large-scale clustering. ARO aims at predicting whether a node should be linked to its $k$ Nearest Neighbors ($k$NN), and transitively merges all linked pairs. Therefore, the computational complexity of ARO is only $O(kn)$. The $k$NN search process can be also accelerated by the Approximate Nearest Neighbor (ANN) search algorithm. Accordingly, the overall complexity is $O(n\log n)$ or $O(n^2)$, depending on whether we set $k$ as a constant or let it increase with $n$. ARO is much more efficient than AHC based algorithms. Shi~\etal~\cite{con} also adopt ANN to expand their ConPac algorithm to a scalable version. In this work, since the proposed method is based on $k$NN, it is suitable to exploit this framework as well.

\textbf{Link Prediction} is a key problem in social network analysis~\cite{linkprediction1, friendlink, movielink,linkprediction2, linkprediction}.
Given a complex network which is organized as a graph, the goal is to predict the likelihood of link between two member nodes. 
To estimate the likelihood of links, some previous works like PageRank~\cite{pagerank} and SimRank~\cite{simrank} analyze the entire graph, while others, such as preferential attachment~\cite{pa} and resource allocation~\cite{ra}, calculate the link likelihood of the given nodes only from their neighbors. 
Zhang and Chen~\cite{wlnm,linkprediction} argue that it is sufficient to compute link likelihood only from the local neighbor of a node pair, and propose a Weisfeiler-Lehman Neural Machine~\cite{wlnm} and a graph neural network~\cite{linkprediction} to learn general graph structure features from local sub-graphs. 
It is closely related to our work, since the clustering task can be reduced to a link prediction problem, and we also exploit graph neural networks to learn from local graphs.

\textbf{Graph convolutional network (GCN).} In many machine learning problems, the input can be organized as graphs. 
Considerable research effort~\cite{spectralgcn1,chebynet,gcn,gat,graphsage} has been devoted to designing the convolutional neural network for graph-structured data. 
According to the definition of convolution on graph data, GCNs can be categorized into spectral methods and spatial methods.
Spectral based GCNs~\cite{spectralgcn1, chebynet, gcn} generalize convolution based on Graph Fourier Transform, while spatial based GCNs~\cite{gat,graphsage} directly perform manually-defined convolution on graph nodes and their neighbors.
In terms of applications, GCNs can handle problems in both the transductive setting~\cite{chebynet,gcn} and the inductive setting~\cite{graphsage,gat}. 
In the transductive setting, training data and testing data are nodes in a same fixed graph, while in the inductive setting, the model needs to inference across different graphs.
In this work, we propose a spatial-based GCN to solve  link prediction problem. The designed GCN performs graph node classification in the inductive setting. 
\section{Proposed Approach}
\label{method}
\subsection{Overview}

\begin{table}[t]
\footnotesize
    \centering
    \begin{tabular}{lcccccc }
    \toprule
         $k$& 5&10 &20& 40& 80 & 160 \\
         \hline
         F-measure& 0.874& 0.911& 0.928& 0.946&0.959&0.970 \\
         NMI& 0.960&0.969&0.975&0.981&0.986&0.990\\
    \bottomrule
    \end{tabular}
    \caption{Upper bound of face clustering accuracy on the IJB-B-512 dataset. We use two metrics for evaluation, F-measure and NMI (see Section \ref{exp:eval}). The upper bound is obtained by connecting each instance with its $k$ nearest neighbors \emph{if the neighbor is of the same identity with this instance.} We find that the upper bound is reasonably high, indicating that $k$NN method could be effective while, most importantly, being efficient.}
    \label{tab:gtlink}
\end{table}

\textbf{Problem definition.} Assume that we have the features of a collection of face images $X = [\bm{x_1}, ..., \bm{x_N}]^{T} \in \mathbb{R}^{N\times D}$, where $N$ is the number of images and $D$ the dimension of the features, the goal of face clustering is to assign a pseudo label $y_i$ to each 
$i \in \{1,2,...,N\}$ so that instances with the same pseudo label form a cluster.
To resolve this problem, we follow the link-based clustering paradigm, which aims at predicting the likelihood of linkage between pairs of instances. 
Accordingly, clusters are formed among all the instances connected by linked pairs. 

\textbf{Motivation.} 
The motivation behind this work is that we find we only need to compute the linkage likelihood between an instance and its $k$ nearest neighbors, and it suffices to produce decent clustering results.
In Table \ref{tab:gtlink}, we show an upper bound of clustering performance with different values of $k$. 
To obtain the upper bound, we directly connect each instance with its $k$NN if the neighbor is of the same identity with this instance. 
The results show that the upper bound is quite high under various values of $k$.
This indicates the potential effectiveness of predicting linkages between an instance and its $k$NN, rather than among all potential pairs.
The advantage of adopting such a strategy is that we could obtain reasonably high clustering accuracy while the system has the benefit of being efficient. 

\begin{figure*}
    \centering
    \includegraphics[width=\linewidth]{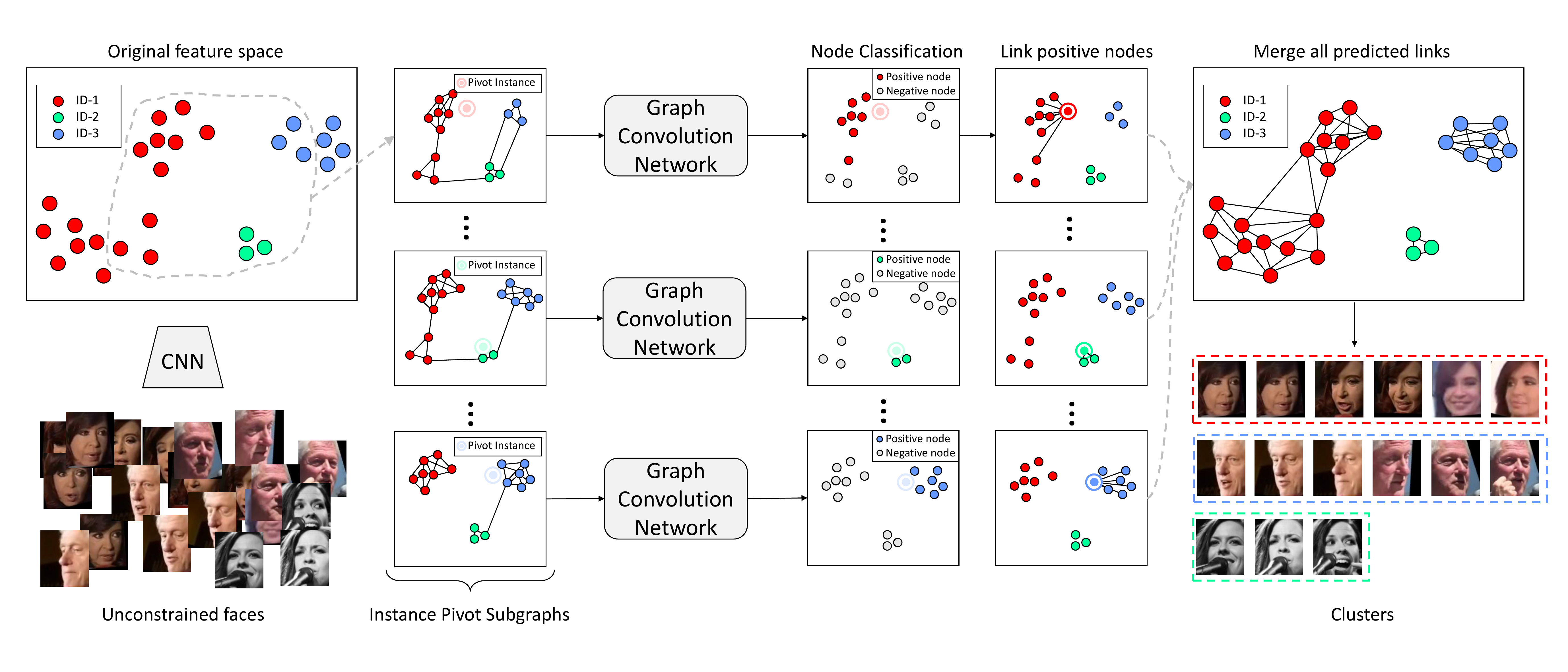}
    \caption{Framework of our method. We use every instance as a pivot, and construct an Instance Pivot Subgraph (IPS) for it. Each node in IPS models the linkage (similarity) between the pivot and the corresponding neighbor. We apply graph convolutions on IPS and classify nodes in IPS into either positive or negative. If a node is classified as positive, the corresponding neighbor should be linked to the pivot. After determining all the linkages, we transitively merge linked instances to obtain the final clusters.}
    \label{fig:pipeline}
\end{figure*}

\textbf{Pipeline.} This work focuses on the efficiency and accuracy of a face clustering system. So we adopt the idea of predicting linkages between an instance and its $k$NNs. 
Because predicting a linkage is based on its context, to make linkage prediction possibly accurate, we design a local structure named Instance Pivot Subgraphs (IPS). 
An IPS is a subgraph centered at a pivot instance $p$. IPS is comprised of nodes including the $k$NNs of $p$ and the high-order neighbors up to $h$-hop of $p$. 
\emph{Importantly, we subtract the feature of pivot $p$ from all these nodes, so that each resulting node feature encodes the linkage relationship between a pivot-neighbor pair.}
We present the framework of the proposed approach in the following three steps, and an illustration is shown in Fig. \ref{fig:pipeline}:
\begin{itemize}
\item  We use every instance as a pivot, and construct an Instance Pivot Subgraph (IPS) for it. 
The construction of IPS is described in detail in Section \ref{method:ips}.
\item Given an IPS as input data, we apply graph convolution networks (GCNs) for reasoning on it and the network outputs a score for every node, \ie, linkage likelihood between the corresponding pivot-neighbor pair. 
The mechanism of GCN is presented in Section \ref{method:gcn}.
\item The above steps output a set of weighted edges among the entire graph, where the weights are the linkage likelihood. 
Finally we transitively merge linked instances into clusters,  according to the linkage likelihood. Details are presented in Section \ref{method:merge}.
\end{itemize}


\begin{figure}
    \centering
    \includegraphics[width=\linewidth]{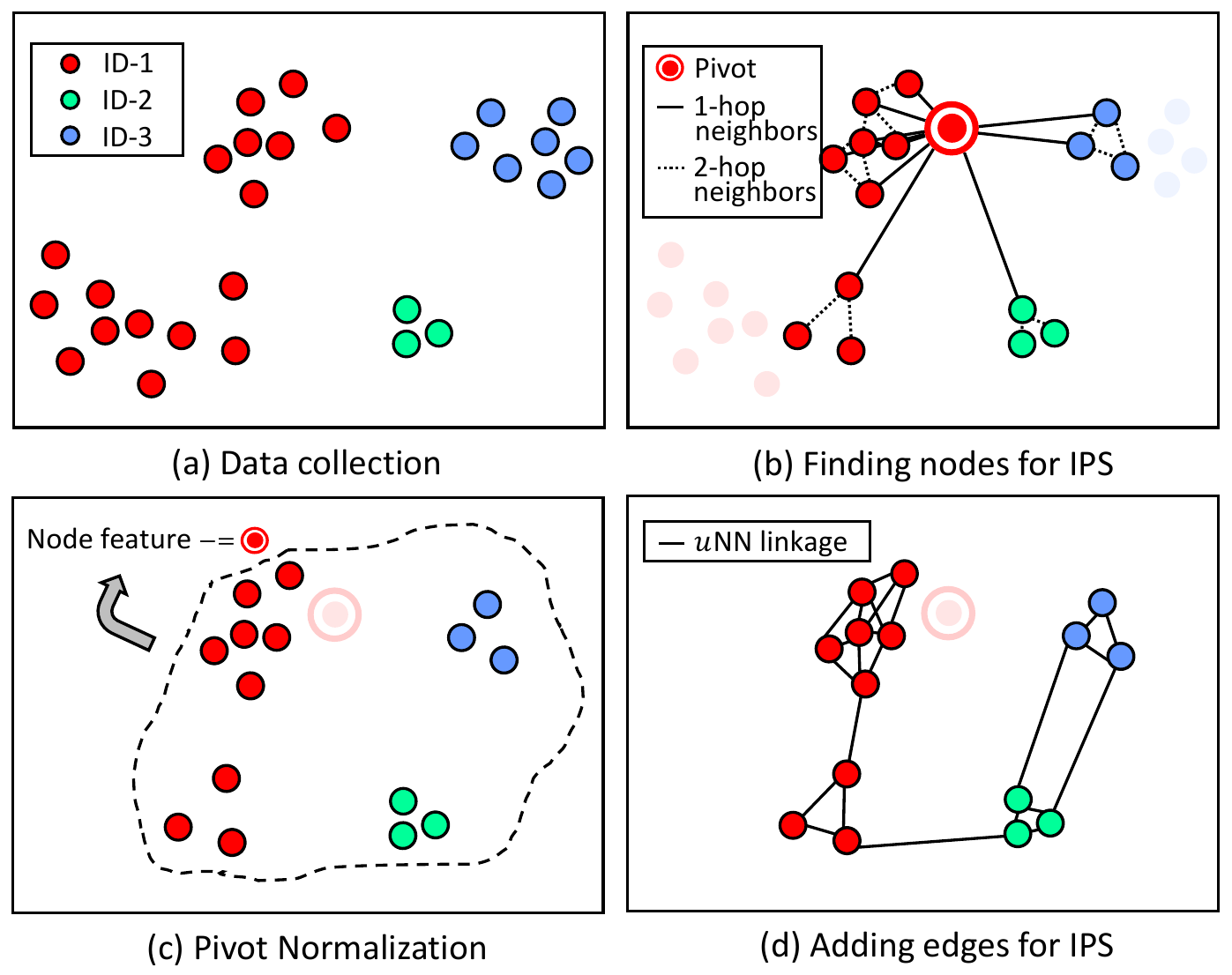}
    \caption{Construction of Instance Pivot Subgraph (IPS). (a) The collection of face representations. (b) We use each instance $p$ as a pivot, and find its neighbors up to $h$-hop as the nodes of IPS. (c) These node features are normalized by subtracting the feature of the pivot. (d) For a node in IPS, we find its $u$NNs from the entire collection. We add an edge between a node and its $u$NNs if the neighbor is also a node of IPS. In this figure, we set $h=2, k_1=10, k_2=2$ and $u=3$, where $k_1$ is the number of 1-hop neighbors, and $k_2$ is the number of of 2-hop neighbors. Note that an IPS based on pivot $p$ does not contain $p$. The IPS for pivot $p$ is used to predict the linkage between $p$ and every node in IPS.}
    \label{fig:ips}
\end{figure}


\subsection{Construction of Instance Pivot Subgraph}
\label{method:ips}

We estimate the linkage likelihood between two face images (nodes) based on their context in a graph. 
In this paper, we propose to construct the Instance Pivot Subgraph (IPS) as context. IPS is generated by three steps.  
First, we locate all the nodes of IPS. 
Then, we normalize the node features by subtracting the feature of the pivot.
Finally, we add edges among nodes. 
An illustration of Instance Pivot Subgraph generation is presented in Fig. \ref{fig:ips}.

\textbf{Step 1: Node discovery.}
Given a pivot $p$, we use its neighbors up to $h$-hop as nodes for IPS. 
For each hop, the  number of picked nearest neighbors may vary.
We denote the number of nearest neighbors in the $i$-th hop as $k_i, i=1,2,...,h$. 
For example, let $p$ be the pivot, the node set $V_p$ of the IPS $G_p(V_p,E_p)$ with $h=3$ and $k_1=8, k_2=4, k_3=2$ consists of 8 nearest neighbors of $p$, 4 nearest neighbors of each 1-hop neighbors, and 2 nearest neighbors of each two-hop neighbors. Note that the pivot $p$ itself is excluded from  $V_p$.
When we do so, the high-order neighbors provide auxiliary information of the local structure of the context between a pivot and its neighbor. 
For instance, for $p$ and one of its neighbors $q$, if the 
$k$NN of $q$ are consistently far away from $p$, then the linkage likelihood between $p$ and $q$ is expected be small.

\textbf{Step 2: Node feature normalization.}
Now we have the pivot instance $p$, the node set $V_p$ and their node features $\bm{x_p}$ and $\{\bm{x_q}| q\in V_p\}$.
In order to encode the pivot information into the node features of IPS, we normalize the node features by subtracting $\bm{x_p}$,
\begin{equation}
    \bm{\mathcal{F}_p} = 
    [\hdots, \bm{x_q} -\bm{x_p}, \hdots]^T
    , \text{for all $q \in V_p$},
\end{equation}
where we use $\bm{\mathcal{F}_p}\in \mathbb{R}^{|V_p|\times D}$ to represent the normalized node features. A node feature in IPS is the residual vector between the feature of the pivot $p$ and the feature of the corresponding neighbor $q$.

\textbf{Step 3: Adding edges among nodes.}
The last step is to add edges among the nodes. 
For a node $q\in V_p$, we first find the top $u$ nearest neighbors among all instances in the original entire collection.
If a node $r$ of the $u$NNs appears in $V_p$, we add an edge $(q,r)$ into the edge set $E_p$. 
This procedure ensures the degree of nodes does not vary too much. 
Finally we represent the topological structure of IPS by an adjacency matrix $\bm{A_p}\in \mathbb{R}^{|V_p|\times |V_p|}$ and the node feature matrix $\bm{\mathcal{F}_p}$.  We neglect the subscript $p$ 
hereafter.

\subsection{Graph Convolutions on IPS}
\label{method:gcn}
The context  contained in IPS (edges among the nodes) is highly valuable for determining if a node is is positive (should link to the pivot) or negative (should not link to the pivot).  
To leverage it, we apply the graph convolution networks (GCN)~\cite{gcn} with slight modifications to perform reasoning on IPS.
A graph convolution layer takes as input a node feature matrix $\bm{X}$ together with an adjacency matrix $\bm{A}$ and outputs a transformed node feature matrix $\bm{Y}$.
In the first layer, the input node feature matrix is the original node feature matrix, $\bm{X}=\bm{\mathcal{F}}$.
Formally, a graph convolution layer in our case has the following formulation,
\begin{equation}
    \bm{Y} = \sigma([\bm{X}\|\bm{G}\bm{X}]\bm{W}),
\end{equation}
where $\bm{X} \in \mathbb{R}^{N\times d_{in}}$, $\bm{Y} \in \mathbb{R}^{N\times d_{out}}$, $N$ is the number of nodes, and $d_{in}$, $d_{out}$ are the dimension of input / output node features.
$\bm{G} = \bm{g}(\bm{X},\bm{A})$ is an aggregation matrix of size $N \times N$ and each row is summed up to $1$, and $\bm{g(\cdot)}$ is a function of $\bm{X}$ and $\bm{A}$. 
Operator $\|$ represents matrix concatenation along the feature dimension.
$\bm{W}$ is the learnable weight matrix of the graph convolution layer of size $2d_{in} \times d_{out}$, and
$\sigma(\cdot)$ is the non-linear activation function.

The graph convolution operation can be broken down into two steps. In the first step,
by left multiplying $\bm{X}$ by $\bm{G}$, the underlying information of nodes' neighbors is aggregated.
Then, the input node features $\bm{X}$ are concatenated with the aggregated information $\bm{G}\bm{X}$ along the feature dimension.
In the second step, the concatenated features are transformed by a set of linear filters, whose parameter $\bm{W}$ is to be learned.
The following three strategies $\bm{g(\cdot)}$ for the aggregation operation is explored.
\begin{itemize}
    \item \textbf{Mean Aggregation}. The aggregation matrix $\bm{G}=\bm{\Lambda^{-\frac{1}{2}}}\bm{A}\bm{\Lambda^{-\frac{1}{2}}}$, where $\bm{A}$ is the adjacency matrix and $\bm{\Lambda}$ is a diagonal matrix with $\bm{\Lambda}_{ii}=\sum_j \bm{A}_{ij}$. 
    The mean aggregation performs average pooling among neighbors.
    \vspace{-0.6cm}
    \item \textbf{Weighted Aggregation}.
    We replace each non-zero element in $\bm{A}$ with the corresponding cosine similarity, and use softmax function to normalize these non-zero values along each row. 
    The weighted aggregation performs weighted average pooling among neighbors.
    \vspace{-0.2cm}
    \item \textbf{Attention Aggregation}. Similar to the graph attention network~\cite{gat}, we attempt to \textit{learn} the aggregation weights of neighbors. That is, the elements in $\bm{G}$ are generated by a two-layer MLP using the features of a pair of pivot-neighbor nodes as input. The MLP is trained end-to-end. The attention aggregation performs weighted average pooling among neighbors, where the weights are learned automatically.
\end{itemize}

The GCN used in this paper is the stack of four graph convolution layers activated by the ReLU function. 
We then use the cross-entropy loss after the softmax activation as the objective function for optimization.
In practice, we only backpropagate the gradient for the nodes of the 1-hop neighbors, because we only consider the linkage between a pivot and its 1-hop neighbors. 
This strategy not only leads to considerable acceleration compared with the fully supervised case, but  also brings about accuracy gain. The reason is that the high-order neighbors are mostly negative, so the positive and negative samples are more balanced in 1-hop neighbors than in all the neighbors.
For testing, we only perform node classification on the 1-hop nodes as well.

\begin{figure}
    \centering
    \includegraphics[width=0.25\linewidth]{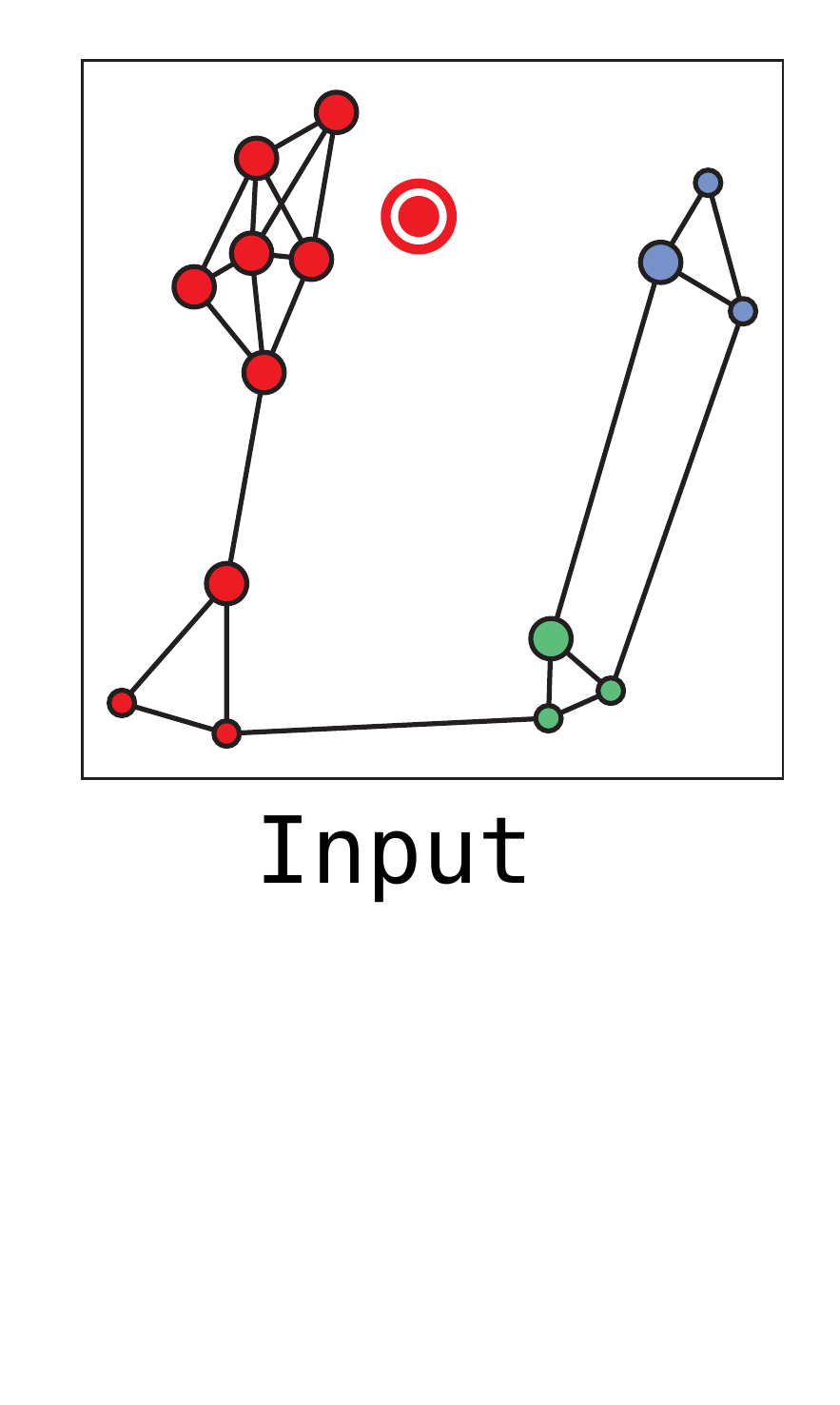}
    \includegraphics[width=0.7\linewidth]{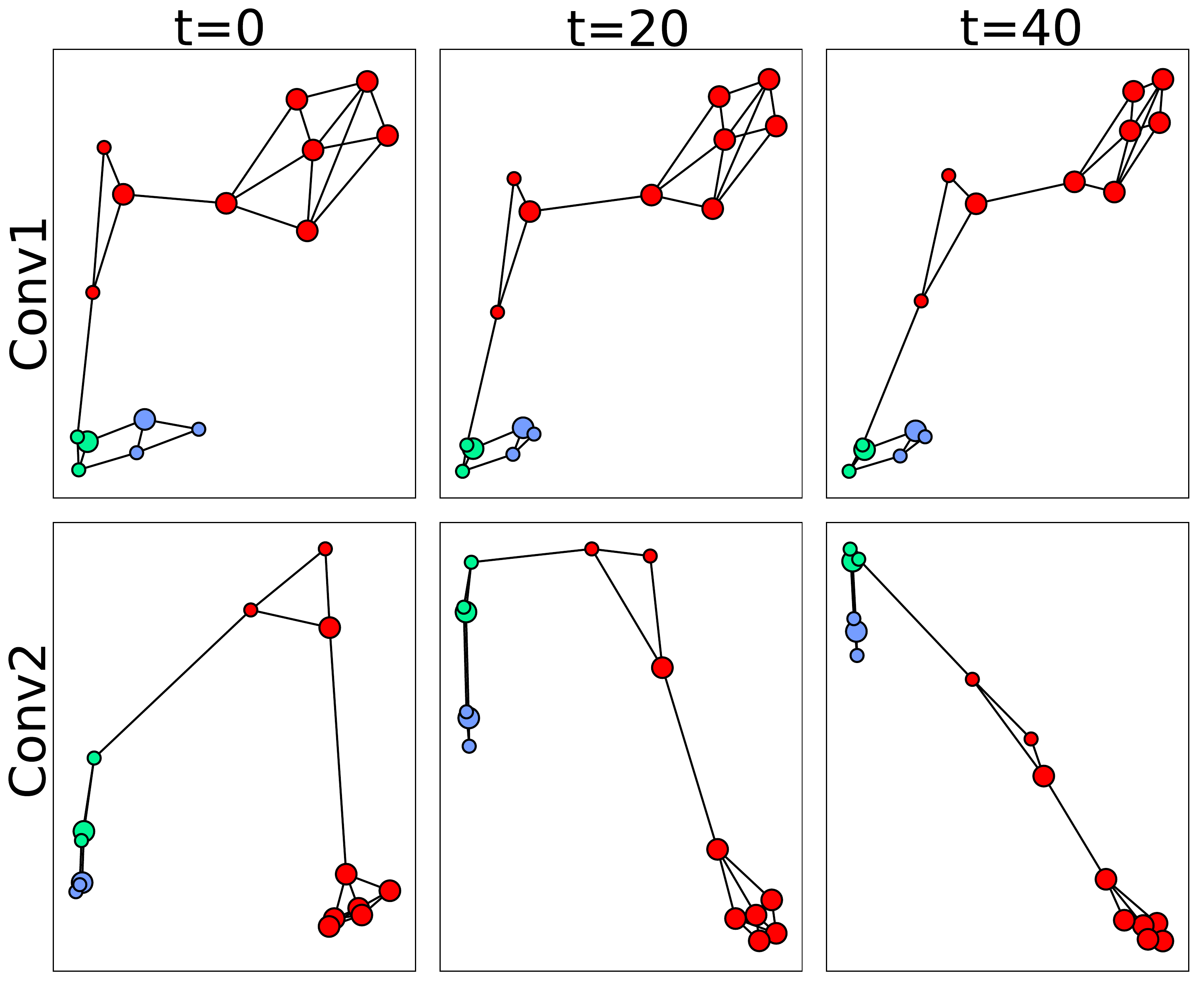}
    \caption{A toy example to illustrate the working mechanism of graph convolutions. Different colors refer to different IDs. The pivot is circled. Training gradients are bac-kpropagated for 1-hop neighbors (larger nodes) but not for higher-order neighbors (smaller nodes). We observe that, after each graph convolution, positive  and  negative  groups become farther from each other and nodes in the same category become closer .}
    \label{fig:gcnprocess}
    
\end{figure}

To demonstrate the working mechanism of graph convolutions, we design a toy example with 2-D input node feature and two graph convolution layers.
The output dimension of each layer, $d_1$ and $d_2$, is set to 2 for visualization purpose.
In Fig. \ref{fig:gcnprocess}, we show how the output embeddings in each layer vary with training iterations. 
After each graph convolution layer, positive nodes (red) are grouped closer, and the negative nodes (blue and green) form another group. This is because messages of neighbors are passed to nodes in the aggregation step, and the message from neighbors acts as a smoothness for the embedding that pulls linked nodes together, like these nodes are connected by springs. Meanwhile, the supervision signal pushes away the group of positive nodes and the group of negative nodes. 
Finally, the system reaches its equilibrium point, where positive and negative groups are far from each other and nodes in the same category are close.

\subsection{Link Merging}
\label{method:merge}
To apply clustering on a collection of face data, we loop over all the instances, construct an IPS with each instance as the pivot, and predict the likelihood of linkage (the softmax probability output by the node classifier) between the involved instances and the pivot.
As a result, we have a set of edges weighted by the linkage likelihood. 
To acquire clusters, a simple approach is to cut all the edges whose weight is below some threshold and use Breath First Search (BFS) to propagate pseudo labels, as shown in Fig. \ref{fig:pipeline}. 
However, the performance can be sensitively affected by the threshold value. We accordingly adopt the pseudo label propagation strategy proposed in \cite{cdp}. 
In each iteration, the algorithm cuts edges below some threshold and maintain connected clusters whose size are larger than a pre-defined maximum size in a queue to be processed in the next iteration.
In the next iteration, the threshold for cutting edges is increased. This process is iterated until the queue is empty, which means all the instances are labeled with pseudo labels.

\section{Experiment}
\label{experiments}

\subsection{Evaluation Metrics and Datasets}

\label{exp:eval}
To evaluate the performance of the proposed clustering algorithm, we adopt two mainstream evaluation metrics: normalized mutual information (NMI) and BCubed F-measure~\cite{bcubed}. 
Given $\bm{\Omega}$ the ground truth cluster set, $\bm{C}$ the predicted cluster set, NMI is defined as, 
\begin{equation}
    NMI(\bm{\Omega}, \bm{C}) = \frac{I(\bm{\Omega}, \bm{C})}{\sqrt{H(\bm{\Omega})H(\bm{C})}},
\end{equation}
where $H(\bm{\Omega})$ and $H(\bm{C})$ represent the entropies for $\bm{\Omega}$ and $\bm{C}$, and $I(\bm{\Omega},\bm{C})$ is the mutual information. 

BCubed F-measure~\cite{bcubed} is a more practical measurement which takes both precision and recall into consideration. 
Let us denote $L(i)$ and $C(i)$ as the ground truth label and cluster label, respectively, we first define the pairwise correctness as, 
\begin{equation}
    Correct(i,j) = 
    \left\{
    \begin{array}{ll}
         1,&\text{if $L(i)=L(j)$ and $C(i)=C(j)$}\\
         0,& \text{otherwise} 
    \end{array}.
    \right.
\end{equation}
The BCubed Precision $P$ and BCubed Recall $R$  are respectively defined as,
\begin{equation}
    P = \mathbb{E}_i[\mathbb{E}_{j:C(j)=C(i)}[Correct(i,j)]],
\end{equation}
\begin{equation}
    R = \mathbb{E}_i[\mathbb{E}_{j:L(j)=L(i)}[Correct(i,j)]],
\end{equation}
and the BCubed F-measure is defined as $F = \frac{2PR}{P+R}$.

We use separate datasets for training and testing. First,  we use   ArcFace~\cite{arcface}\footnote{\href{https://github.com/deepinsight/insightface}{https://github.com/deepinsight/insightface}} as the face representations. This model is trained on the union set of MS-Celeb-1M~\cite{ms1m}  and VGGFace2~\cite{vgg2} dataset. Second, for GCN training, we use a random subset of the CASIA dataset~\cite{casia} which contains ~5k identities and 200k samples. Third, for testing, we adopt the IJB-B dataset~\cite{ijbb} because it contains a clustering protocol. The protocol consists of seven subtasks varying in the number of ground truth identities. 
We evaluate our algorithm on three largest subtasks. In the three subtasks, the numbers of identities are 512, 1,024 and 1,845, and the numbers of samples are 18,171, 36,575 and 68,195, respectively.



\subsection{Parameter Selection}
\begin{figure}
    \centering
    \includegraphics[width=\linewidth]{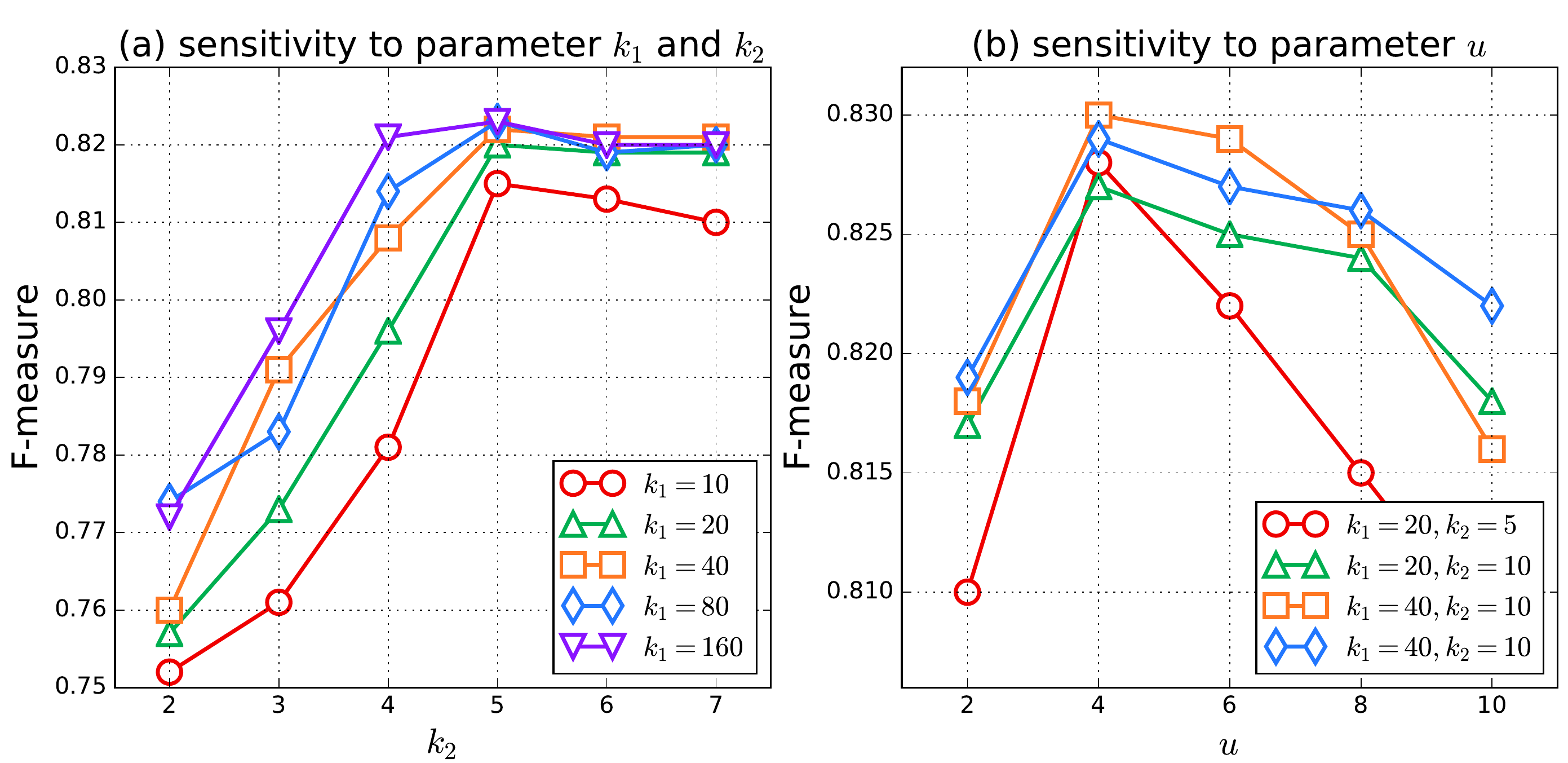}
    \caption{(a) The sensitivity of F-measure on IJB-B-512 to $k_1$ and $k_2$, with constant $u=10$. (b) The sensitivity of F-measure on to parameter $u$, with different combination of $k_1$ and $k_2$.}
    \label{fig:parameters}
    
\end{figure}
There are three hyperparameters for IPS construction: the number of hops $h$, the number of picked nearest neighbors in each hop $\{k_i\}, i=1,2,...,h$, and the number of linked nearest neighbors $u$ for picking edges. 
We first experiment with different values of $h$ and find that $h\geq3$ does not bring performance gain, so we set $h=2$ in the following experiment.
Accordingly, we explore the impact of different values $k_i, k_2$ and $u$.
We discuss both the training phase and the testing phase.

In the training phase, we expect more supervision signals to be back-propagated.
Since supervision is only added to 1-hop nodes, we select a large  $k_1=200$.
In order to avoid IPS being too large, we set a small value $k_2=10$. 
We also set $u=10$ to ensure every 2-hop node has at least one edge.

In the testing phase, it is not necessary to keep the same configuration with the training phase. To investigate how  $k_1, k_2$ and $u$ influence the performance, we conduct two group of experiments and the the results are shown in Fig.~\ref{fig:parameters}. 
First, we keep $u$ constant, vary $k_1$, $k_2$, and show how F-measure changes on 1JB-B-512. 
We observe in Fig.~\ref{fig:parameters} (a) that the F-measure increases with larger $k_1$ and $k_2$.
Larger $k_1$ brings more candidate links to be predicted, thus yields higher recall.
Larger $k_2$ involves more 2-hop neighbors, depicting the local structure of 1-hop neighbors more precisely, so the prediction is more accurate.
However, the performance reaches saturation when $k_1$ and $k_2$ are large enough.
For the parameter $u$, \ie, the linked number of neighbors, we observe in \ref{fig:parameters} (b) that the performance is not sensitive to the value of $u$.

Taking efficiency into consideration, the values of $k_1$ and $k_2$ cannot be too large. 
We find that $k_1=80, k_2=5, u=5$ yield a good trade-off between efficiency and performance and use this setting in the following experiment. 

\subsection{Evaluation}
The proposed approach is compared to the following methods: K-means~\cite{kmeans}, Spectral Clustering~\cite{spectral1}, Agglomerative Hierarchical Clustering (AHC)~\cite{agglomerative1}, Affinity Propagation (AP)~\cite{ap}, Density-Based Spatial Clustering of Applications with Noise (DBSCAN)~\cite{dbscan}, Proximity-Aware Hierarchical Clustering (PAHC)~\cite{pahc}, Deep Density Clustering (DDC)~\cite{ddc}, Conditional Pair-wise Clustering (ConPaC)~\cite{con}, and Approximate Rank-Order Clustering (ARO)~\cite{aro}.
For all the methods, we tune the hyperparameters \eg, $\sigma$ in Spectral Clustering and $\epsilon$, $n$ in DBSCAN , 
and report the best results.
For non-deterministic algorithms like K-means we select the best result from 100 runs. 

\textbf{Comparing different aggregation methods.} 
We first compare the aggregation strategies described in Section \ref{method:gcn}. 
In Table~\ref{tab:comparison}, GCN-M refers to Mean Aggregation, GCN-W refers to Weighted Aggregation, and GCN-A refers to Attention Aggregation. 
The Attention Aggregation learns the aggregation weights of neighbors automatically in an end-to-end manner, yielding marginally better performance than Mean Aggregation and Weighted Aggregation. 
Considering the computation cost, the improvement is not significant, so we use Mean Aggregation in the following experiment.

\textbf{Comparison with baseline methods.} 
The top part of Table \ref{tab:comparison} showcases results of several widely used clustering algorithms. 
The results suggest algorithms that make less restrictive assumptions on  data distribution usually achieve higher performance. 
For instance, there is no assumptions on data distribution in AHC and the performance is the best.
DBSCAN requires the data to have similar density and the performance is inferior to AHC. 
K-means needs the data to be convex-shaped, and Spectral Clustering is not good at handling unbalanced data, thus both yield unsatisfactory results. 
Same as AHC, our approach does not make any assumptions on the data distribution, and the clustering rule is learned by a parametric model, therefore it is not surprising the performance is superior to the strong AHC baseline.
This is  not a trivial result, since the performance of AHC is sensitive to the threshold, while ours is not sensitive to parameter selection and consistently outperforms AHC.

\begin{table}[t]
\footnotesize
    \centering
    \begin{tabular}{lcccccc}
    \toprule
         \multirow{2}{*}{Method}& \multicolumn{2}{c}{IJB-B-512}  & \multicolumn{2}{c}{IJB-B-1024} & \multicolumn{2}{c}{IJB-B-1845} \\
    \cline{2-7}
         & F & NMI  & F & NMI  & F & NMI \\

    \hline
    K-means~\cite{kmeans}
    &0.612	& 0.858& 0.603	& 0.865& 0.600& 0.868\\
    Spectral~\cite{spectral1} 
    & 0.517	& 0.784& 0.508	& 0.792&0.516 & 0.785\\
    AHC~\cite{agglomerative1} 
    &0.795 & 0.917 &0.797& 0.925&0.793 & 0.923\\
    AP~\cite{ap} 
    & 0.494	& 0.854&0.484 & 0.864& 0.477& 0.869 \\
    DBSCAN~\cite{dbscan} 
    &0.753 & 0.841 &0.725 & 0.833 &0.695 &	0.814 \\
    \hline
    \hline
    ARO~\cite{aro} 
    &0.763 & 0.898 &0.758 & 0.908 & 0.755 & 0.913\\
    PAHC$^*$~\cite{ddc} 
    &-& - &0.639 & 0.890 & 0.610 & 0.890\\
    ConPaC$^*$~\cite{con} 
    &0.656 & - &0.641 & - &0.634 & -\\
    DDC~\cite{ddc} 
    & 0.802 & 0.921 & 0.805& 0.926 &0.800 &0.929 \\
    \hline
    \hline
    GCN-M
    & 0.821&0.920 & 0.819 & 0.928&0.810	 & 0.927\\    
    GCN-W
    & 0.826& 0.923 & 0.822 & 0.938 	& 0.801 & 0.927 \\    
    GCN-A
    & \textbf{0.833}& \textbf{0.936} &\textbf{ 0.833} & \textbf{0.942}&	\textbf{0.814} & \textbf{0.938} \\ 
    \bottomrule
    
    \end{tabular}
    \caption{Comparison with baseline methods in terms of BCubed F-measure and NMI score. For all methods we tune the corresponding hyperparameters and report the best result.  Suffix M, W, and A represents different aggregators. The superscript~$*$ denotes results reported from the original papers, otherwise all methods use the same ArcFace representation.}
    \label{tab:comparison}
    
\end{table}
\textbf{Comparison with state-of-the-art.} 
In the second part of Table~\ref{tab:comparison} we compare our method with four state-of-the-art face clustering algorithms, \ie, ARO~\cite{aro}, PAHC~\cite{pahc}, ConPaC~\cite{con} and DDC~\cite{ddc}. 
The proposed method consistently outperforms other method on the three subtasks in term of both F-measure and NMI score. 
Note that the results of PAHC and ConPac may not be compared directly since different face representations are employed. However, we find that both of them underperform the corresponding AHC baseline (with the same face representation), while our method surpass the AHC baseline. 
This shows the accuracy of our mehtod is favorably comparable to the state-of-the-art face clustering algorithms.

\textbf{Different face representation.}  To validate that the benefit is indeed from the algorithm rather than the strong ArcFace feature, we train a face recognition model using ResNet-50~\cite{resnet} + Softmax Loss on the MS1M dataset~\cite{ms1m}, and test clustering methods with such representation. Method comparison under Res-50 representation is shown
in Table. \ref{tab:feature}. Combined with Table. \ref{tab:comparison}, the results show that:

(1) When stronger representation is adopted (Arcface),
our method yields better performance. This indicates our
method is able to benefit from better representation.

(2) When using the same representation, our method outperforms state-of-the-art methods. This indicates our method
has superior performance to prior arts.

\begin{table}[t]
\footnotesize
    \centering
    \begin{tabular}{lcccccc}
    \toprule
         \multirow{2}{*}{Method}& \multicolumn{2}{c}{IJB-B-512}  & \multicolumn{2}{c}{IJB-B-1024} & \multicolumn{2}{c}{IJB-B-1845} \\
    \cline{2-7}
         & F & NMI  & F & NMI  & F & NMI \\
         
    \hline
    AHC~\cite{agglomerative1}
    &0.688 & 0.874 &0.694& 0.880&0.676 & 0.867\\
    ARO~\cite{aro} 
    &0.624 & 0.852 &0.628 & 0.853 & 0.619 & 0.848\\
    DDC~\cite{ddc} 
    & 0.704 & 0.887 & 0.708& 0.892 &0.697 &0.889 \\
    Ours
    & \textbf{0.736}& \textbf{0.910} &\textbf{ 0.733} & \textbf{0.913}&	\textbf{0.726} & \textbf{0.908} \\ 
    \bottomrule
    
    \end{tabular}
    \caption{Method comparison under the same Res-50 feature.}
    \label{tab:feature}
    
\end{table}

\textbf{Singleton Clusters.}
\begin{figure}
    \centering
    \includegraphics[width=\linewidth]{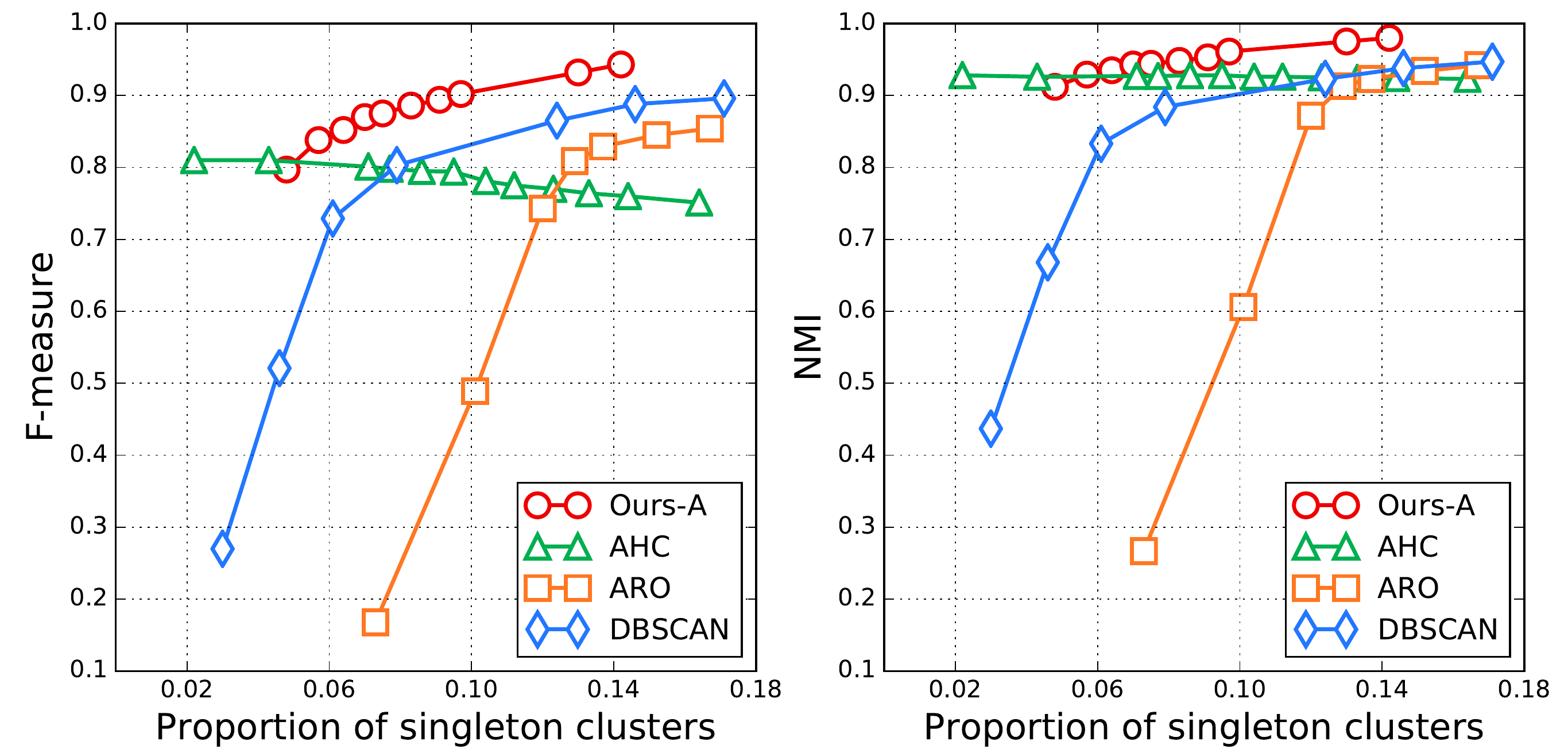}
    \caption{Performance on IJB-B-512 after removing singleton clusters \textit{v.s.} proportion of  removed singleton clusters.}
    \label{fig:singletoncurve}
\end{figure}
In practice, we find our approach produces many singleton clusters, \ie, clusters that contain only a single sample.
The proportion of the generated singleton clusters among the whole collection varies with the hyperparameters in the merging step. 
We examine singleton samples and find that most of the them are extreme hard samples, \ie,  profile faces, low-resolution faces or blurred faces,
also non-face images due to the failure of face detection, and mis-labeled faces.
We filtered all the singleton clusters and re-test the F-measure and NMI score on IJB-B-512.
For a fair comparison, we also report the performance of other three linkage based clustering methods after removing singleton clustering. 
We manually tune the hyperparameters in each algorithm to let  the proportion of singleton instances vary, then remove singleton clusters and compute  F-measure and NMI score.
Finally We plot the curves of the two metrics as the proportion of singleton clusters varies in Fig.~\ref{fig:singletoncurve}.
ARO, DBSCAN and our method present ascending cureves, which means these methods are able to filter out noise and outliers. 
By tuning hyperparameters to be stricter, these methods generate more singleton clusters, and the remained non-singleton clusters are more accurate. 
In contrast, AHC presents a plain curve, so the accuracy of generated clusters is not controllable by tuning hyperparameters.
With the same proportion of singleton clusters, our method consistently outperforms other algorithms. 
Furthermore, clusters generated by our method are in pretty high purity, with a moderate sacrifice in instance number (say 0.943 F-measure with $15.2\%$ instances discarded).
This is a valuable property in applications like automatically data labeling, where the purity is important.

\begin{table}[]
    \centering
    \begin{tabular}{lccp{1.8cm}<{\centering}}
    \toprule
         Parameters&  F-measure &  NMI & runtime\\
         \midrule
         $k_1=10,k_2=5$&  0.634&	0.886&	00:19:50 \\
        $k_1=40,k_2=5$& 0.655	&0.891& 00:53:56\\
         $k_1=160,k_2=5$& 0.720  & 0.905  & 02:45:36\\
    \bottomrule
    \end{tabular}
    \caption{Results on the IJB-B-1845+1M dataset. The total number of images is 1,094,842. Runtime is presented in HH:MM:SS. }
    \label{tab:1M}
    \vspace{-0.3cm}
\end{table}

\textbf{Scalability and Efficiency.} 
The proposed method only operate on local IPS, hence the runtime of link prediction process grows linearly with the number of data. 
The IPS construction has an $O(n^2)$ complexity if we search nearest neighbor by brute force, and can be reduced to $O(n\log n)$ by Approximate Nearest Neighbor (ANN) search.
In general, the overall complexity of our method is $O(n\log n)$, which means it is efficient and scalable.
Here we perform a large-scale experiment by introducing 1 million distractors to the IJB-B-1845 dataset, to investigate the efficiency of our method. 
We use the whole Megaface~\cite{megaface} dataset as distractors, which comprises of 1 million irrelevant face images. 
F-measure and NMI score are then computed by simply ignoring the distractors.
As shown in Table \ref{tab:1M},  the runtime and performance is influenced by $k_1$, and we can tune the $k_1$ for accuracy-time trade-off. All the experiment are performed on a single Titan Xp GPU, and one can use more for acceleration since our algorithm is suitable for parallelism.

\subsection{Multi-View Extension}
In many real-world applications, data may come from multiple sources and contain complementary information, known as "multi-view data". 
Multi-View clustering aims at exploiting such data to generate better clusters. 
In this section, we show our clustering method is easily extended to a multi-view version, and also adaptive to different base features.

We apply the proposed clustering method to video face clustering task, where two views of data, namely face features and audio features, can be extracted to depict a person. 
The face features and the audio features are extracted by two CNNs and then simply concatenated as a joint representation, accordingly the training and inference procedures of the GCN are the same as described above. 

We adopt VoxCeleb2~\cite{vox2} dataset for training the GCN and evaluating the clustering results. 
The VoxCeleb2 dataset comprises of 145K videos of 5,994 different identities, and we split it into a test set consisting of 2,048 identities and a disjoint training set.
We propose two clustering protocols which consist of 22,568 instances of 512 identities and 83,265 instances of 2,048 identities, respectively. 
Several clustering methods are compared with three different base features, namely face, audio and face+audio, and the results are presented in Table \ref{tab:multi512} and Table \ref{tab:multi2048}.
\begin{table}[]
\footnotesize
    \centering
    \begin{tabular}{lcccccc}
    \toprule
         \multirow{2}{*}{Method}& \multicolumn{2}{c}{Face}  & \multicolumn{2}{c}{Audio} & \multicolumn{2}{c}{Face+Audio} \\
    \cline{2-7}
         & F & NMI  & F & NMI  & F & NMI \\

    \hline
    K-means~\cite{kmeans}
    &0.648	& 0.877& 0.229	& 0.644& 0.636& 0.874\\
    Spectral~\cite{spectral1} 
    & 0.592	& 0.825& 0.214	& 0.619&0.541 & 0.782\\
    AHC~\cite{agglomerative1} 
    &0.755 & 0.913 &0.358& 0.704&0.833 & 0.934\\
    ARO~\cite{aro} 
    & 0.575 & 0.875 &0.261 & \textbf{0.834} &0.319 &	0.835 \\
    Ours
    &\textbf{0.801}&\textbf{0.921}&\textbf{0.395}&0.497&\textbf{0.841}&\textbf{0.940} \\
     
    \bottomrule
    
    \end{tabular}
    \caption{Clustering accuracy with 512 identities.}
    \label{tab:multi512}
\end{table}

\begin{table}[]
\footnotesize
    \centering
    \begin{tabular}{lcccccc}
    \toprule
         \multirow{2}{*}{Method}& \multicolumn{2}{c}{Face}  & \multicolumn{2}{c}{Audio} & \multicolumn{2}{c}{Face+Audio} \\
    \cline{2-7}
         & F & NMI  & F & NMI  & F & NMI \\

    \hline
    K-means~\cite{kmeans}
    &0.589	& 0.871& 0.152	& 0.650& 0.582& 0.871\\
    AHC~\cite{agglomerative1} 
    &0.695 & 0.908 &0.228 & 0.686 &0.785 &	0.938 \\
    ARO~\cite{aro} 
    &0.583 & 0.858 &0.277& \textbf{0.813}&0.370 & 0.873\\

    Ours
    &\textbf{0.766}&\textbf{0.932}&\textbf{0.311}&0.452&\textbf{0.810}&\textbf{0.946} \\
    \bottomrule
    
    \end{tabular}
    \caption{Clustering accuracy with 2,048 identities.}
    \label{tab:multi2048}
     \vspace{-0.3cm}
\end{table}
The distribution of the concatenated face+audio features are more complex than the single face / audio features, therefore some heuristic clustering methods fail to leverage the complementary information (face features outperform face+audio features). 
In contrast, the proposed method learn the clustering rule by a parametric model, thus is able to handle such data distribution, and brings about performance gain from multi-view data.
This series of experiment show 
our clustering method can be 1) easily extend to a multi-view version, only if training data is provided, and 
also 2) adaptive to different base features.

\section{Conclusion}
\label{conclusion}
In this paper, we propose a linkage based  method for face clustering. 
We emphasize the importance of \emph{context} in face clutering and propose to construct instance pivot subgraphs (IPS) that depict the context of given nodes. On IPS, We use the graph convolution network to reason the linkage likelihood between a given node and its neighbors.
Extensive experiment indicates the proposed method is more robust to the complex distribution of faces than conventional methods. We report favorably comparable results to state-of-the-art methods on standard face clustering benchmarks and show our method is scalable to large datasets. Finally, we demonstrate the strength of our method in visual-audio face clustering. 

\section{Acknowledgement}
This work was supported by the state key development program in 13th Five-Year under Grant No. 2016YFB0801301 and the National Natural Science Foundation of China under Grant Nos. 61771288、61701277.

{\footnotesize
\bibliographystyle{ieee}
\bibliography{egbib}

\begin{thebibliography}{10}\itemsep=-1pt

\bibitem{friendlink}
L.~A. Adamic and E.~Adar.
\newblock Friends and neighbors on the web.
\newblock {\em Social networks}, 25(3):211--230, 2003.

\bibitem{bcubed}
E.~Amig{\'o}, J.~Gonzalo, J.~Artiles, and F.~Verdejo.
\newblock A comparison of extrinsic clustering evaluation metrics based on
  formal constraints.
\newblock {\em Information retrieval}, 12(4):461--486, 2009.

\bibitem{pa}
A.-L. Barab{\'a}si and R.~Albert.
\newblock Emergence of scaling in random networks.
\newblock {\em science}, 286(5439):509--512, 1999.

\bibitem{pagerank}
S.~Brin and L.~Page.
\newblock The anatomy of a large-scale hypertextual web search engine.
\newblock {\em Computer networks and ISDN systems}, 30(1-7):107--117, 1998.

\bibitem{spectralgcn1}
J.~Bruna, W.~Zaremba, A.~Szlam, and Y.~LeCun.
\newblock Spectral networks and locally connected networks on graphs.
\newblock {\em arXiv preprint arXiv:1312.6203}, 2013.

\bibitem{vgg2}
Q.~Cao, L.~Shen, W.~Xie, O.~M. Parkhi, and A.~Zisserman.
\newblock Vggface2: A dataset for recognising faces across pose and age.
\newblock In {\em IEEE International Conference on Automatic Face \& Gesture
  Recognition (FG),}, 2018.

\bibitem{vox2}
J.~S. Chung, A.~Nagrani, and A.~Zisserman.
\newblock Voxceleb2: Deep speaker recognition.
\newblock In {\em INTERSPEECH}, 2018.

\bibitem{chebynet}
M.~Defferrard, X.~Bresson, and P.~Vandergheynst.
\newblock Convolutional neural networks on graphs with fast localized spectral
  filtering.
\newblock In {\em NIPS}, 2016.

\bibitem{arcface}
J.~Deng, J.~Guo, and S.~Zafeiriou.
\newblock Arcface: Additive angular margin loss for deep face recognition.
\newblock {\em arXiv preprint arXiv:1801.07698}, 2018.

\bibitem{dbscan}
M.~Ester, H.-P. Kriegel, J.~Sander, X.~Xu, et~al.
\newblock A density-based algorithm for discovering clusters in large spatial
  databases with noise.
\newblock In {\em KDD}, 1996.

\bibitem{ap}
B.~J. Frey and D.~Dueck.
\newblock Clustering by passing messages between data points.
\newblock {\em science}, 315(5814):972--976, 2007.

\bibitem{ms1m}
Y.~Guo, L.~Zhang, Y.~Hu, X.~He, and J.~Gao.
\newblock Ms-celeb-1m: A dataset and benchmark for large-scale face
  recognition.
\newblock In {\em ECCV}, 2016.

\bibitem{graphsage}
W.~Hamilton, Z.~Ying, and J.~Leskovec.
\newblock Inductive representation learning on large graphs.
\newblock In {\em NIPS}, 2017.

\bibitem{resnet}
K.~He, X.~Zhang, S.~Ren, and J.~Sun.
\newblock Deep residual learning for image recognition.
\newblock In {\em CVPR}, 2016.

\bibitem{agglomerative1}
A.~K. Jain and R.~C. Dubes.
\newblock Algorithms for clustering data.
\newblock 1988.

\bibitem{simrank}
G.~Jeh and J.~Widom.
\newblock Simrank: a measure of structural-context similarity.
\newblock In {\em Proceedings of the eighth ACM SIGKDD international conference
  on Knowledge discovery and data mining}, pages 538--543. ACM, 2002.

\bibitem{megaface}
I.~Kemelmacher-Shlizerman, S.~M. Seitz, D.~Miller, and E.~Brossard.
\newblock The megaface benchmark: 1 million faces for recognition at scale.
\newblock In {\em CVPR}, 2016.

\bibitem{gcn}
T.~N. Kipf and M.~Welling.
\newblock Semi-supervised classification with graph convolutional networks.
\newblock 2017.

\bibitem{boston}
J.~C. Klontz and A.~K. Jain.
\newblock A case study of automated face recognition: The boston marathon
  bombings suspects.
\newblock {\em Computer}, 46(11):91--94, 2013.

\bibitem{movielink}
Y.~Koren, R.~Bell, and C.~Volinsky.
\newblock Matrix factorization techniques for recommender systems.
\newblock {\em Computer}, (8):30--37, 2009.

\bibitem{linkprediction1}
D.~Liben-Nowell and J.~Kleinberg.
\newblock The link-prediction problem for social networks.
\newblock {\em Journal of the American society for information science and
  technology}, 58(7):1019--1031, 2007.

\bibitem{ddc}
W.-A. Lin, J.-C. Chen, C.~D. Castillo, and R.~Chellappa.
\newblock Deep density clustering of unconstrained faces.
\newblock In {\em CVPR}, 2018.

\bibitem{pahc}
W.-A. Lin, J.-C. Chen, and R.~Chellappa.
\newblock A proximity-aware hierarchical clustering of faces.
\newblock In {\em IEEE International Conference on Automatic Face \& Gesture
  Recognition (FG)}, 2017.

\bibitem{kmeans}
S.~Lloyd.
\newblock Least squares quantization in pcm.
\newblock {\em IEEE transactions on information theory}, 28(2):129--137, 1982.

\bibitem{linkprediction2}
L.~L{\"u} and T.~Zhou.
\newblock Link prediction in complex networks: A survey.
\newblock {\em Physica A: statistical mechanics and its applications},
  390(6):1150--1170, 2011.

\bibitem{mf2}
A.~Nech and I.~Kemelmacher-Shlizerman.
\newblock Level playing field for million scale face recognition.
\newblock In {\em CVPR}, 2017.

\bibitem{aro}
C.~Otto, D.~Wang, and A.~K. Jain.
\newblock Clustering millions of faces by identity.
\newblock {\em IEEE transactions on pattern analysis and machine intelligence},
  40(2):289--303, 2018.

\bibitem{spectral1}
J.~Shi and J.~Malik.
\newblock Normalized cuts and image segmentation.
\newblock {\em IEEE Transactions on pattern analysis and machine intelligence},
  22(8):888--905, 2000.

\bibitem{con}
Y.~Shi, C.~Otto, and A.~K. Jain.
\newblock Face clustering: representation and pairwise constraints.
\newblock {\em IEEE Transactions on Information Forensics and Security},
  13(7):1626--1640, 2018.

\bibitem{svdd}
D.~M. Tax and R.~P. Duin.
\newblock Support vector domain description.
\newblock {\em Pattern recognition letters}, 20(11-13):1191--1199, 1999.

\bibitem{gat}
P.~Velickovic, G.~Cucurull, A.~Casanova, A.~Romero, P.~Lio, and Y.~Bengio.
\newblock Graph attention networks.
\newblock 2018.

\bibitem{ijbb}
C.~Whitelam, E.~Taborsky, A.~Blanton, B.~Maze, J.~C. Adams, T.~Miller, N.~D.
  Kalka, A.~K. Jain, J.~A. Duncan, K.~Allen, et~al.
\newblock Iarpa janus benchmark-b face dataset.
\newblock In {\em CVPR Workshops}, pages 592--600, 2017.

\bibitem{casia}
D.~Yi, Z.~Lei, S.~Liao, and S.~Z. Li.
\newblock Learning face representation from scratch.
\newblock {\em arXiv preprint arXiv:1411.7923}, 2014.

\bibitem{cdp}
X.~Zhan, Z.~Liu, J.~Yan, D.~Lin, and C.~Change~Loy.
\newblock Consensus-driven propagation in massive unlabeled data for face
  recognition.
\newblock In {\em ECCV}, 2018.

\bibitem{wlnm}
M.~Zhang and Y.~Chen.
\newblock Weisfeiler-lehman neural machine for link prediction.
\newblock In {\em Proceedings of the 23rd ACM SIGKDD International Conference
  on Knowledge Discovery and Data Mining}, 2017.

\bibitem{linkprediction}
M.~Zhang and Y.~Chen.
\newblock Link prediction based on graph neural networks.
\newblock {\em arXiv preprint arXiv:1802.09691}, 2018.

\bibitem{ra}
T.~Zhou, L.~L{\"u}, and Y.-C. Zhang.
\newblock Predicting missing links via local information.
\newblock {\em The European Physical Journal B}, 71(4):623--630, 2009.

\bibitem{rankorder}
C.~Zhu, F.~Wen, and J.~Sun.
\newblock A rank-order distance based clustering algorithm for face tagging.
\newblock In {\em CVPR}, 2011.

\end{thebibliography}
}

\end{document}